\documentclass[conference, letterpaper, 10 pt]{ieeeconf}  

\IEEEoverridecommandlockouts                              
\usepackage[utf8]{inputenc}
\usepackage[T1]{fontenc}
\usepackage{amsmath}
\usepackage{leftidx}
\usepackage{xcolor}

\newcommand{\eqtext}[1]{\textrm{\tiny{#1}}}
\usepackage{amsfonts}
\usepackage{subfigure}
\usepackage{graphicx}
\usepackage{algorithm}
\usepackage[noend]{algpseudocode}
\usepackage{mdwtab}
\usepackage{eqparbox}
\usepackage{url}
\usepackage{amsmath,amsfonts,amssymb}
\usepackage{graphicx}
\usepackage[colorlinks=true, allcolors=blue]{hyperref}
\usepackage{xcolor}
\usepackage{algorithm}
\usepackage{arevmath}     
\usepackage[noend]{algpseudocode}
\usepackage{mathtools}
\usepackage{array}
\usepackage{multirow}
\usepackage{amsfonts}
\usepackage{booktabs}
\usepackage{siunitx}
\usepackage{textgreek}
\usepackage{amsmath,amssymb}
\usepackage{mathptmx}
\usepackage{comment}
\title{\LARGE \bf
Learn Proportional Derivative Controllable Latent Space from Pixels
}

\author{Weiyao Wang$^{1}$, Marin Kobilarov$^{2}$, and Gregory D. Hager$^{1}$
\thanks{This work was supported by the U.S. National Science Foundation grants IIS-1900952 to Johns Hopkins University}
\thanks{$^{1}$W. Wang and G. D. Hager are with Computer Science Department,
        Johns Hopkins University, Baltimore, MD 21218, USA (e-mail: wwang121@jhu.edu; hager@cs.jhu.edu).
        }%
\thanks{$^{2}$M. Kobilarov is with Mechanical Engineering Department, Johns Hopkins University, Baltimore, MD 21218, USA (e-mail: marin@jhu.edu).
        }%
}

\begin{document}

\maketitle
\thispagestyle{empty}
\pagestyle{empty}

\begin{abstract}
Recent advances in latent space dynamics model from pixels show promising progress in vision-based model predictive control (MPC). However, executing MPC in real time can be challenging due to its intensive computational cost in each timestep. We propose to introduce additional learning objectives to enforce that the learned latent space is proportional derivative controllable. In execution time, the simple PD-controller can be applied directly to the latent space encoded from pixels, to produce simple and effective control to systems with visual observations. We show that our method outperforms baseline methods to produce robust goal reaching and trajectory tracking in various environments.

\end{abstract}

\section{INTRODUCTION}
    Vision-based control is important to many robotic application where the robot configuration (e.g., joint angles) cannot be directly captured by sensors (e.g., encoders) but an image is available.  Scaling  existing optimal control algorithms to high-dimensional, non-linear vision inputs remains an open challenge. A recent approach, known as Learning Controllable Embedding (LCE), is to first embed the high-dimensional visual input into a low-dimensional latent space, and then to jointly learn a transition dynamics model in the latent space~\cite{watter2015embed, finn2016deep, zhang2019solar, banijamali2018robust, pc3, Levine2020Prediction}. After obtaining such a latent space and the corresponding dynamics model, it is feasible to apply traditional planning algorithms or model-predictive control (MPC)~\cite{camacho2013model,levine2014learning,hirose2019deep,finn2017deep,lynch2020learning}.
    
    Such MPC-based approaches are computationally expensive to execute, and the resulting models are often more complex than the true underlying physical state space. Our key insight is that, even though the visual perception is complex to model, the underlying physical system is often quite simple and is often possible to control with straightforward linear methods such as simple proportional derivative control (PD-control)~\cite{johnson2005pid}. However, the latent space produced by LCE approaches is typically complex and is not amenable to PD-control. The main reason is that both the encoding and the transition dynamics are modeled by nonlinear functions which are not constrained to relate each controllable degree of freedom to an embedding dimension in the simplest possible way. 
    
    In this work, we build on LCE to develop embedding that allow simple PD-control on the latent space. Essentially, our goal is to learn a state embedding function $z = E(x)$ from raw observations (images) $x$ such that taking an action produced by PD-control using the current embedding $z$ and its estimated velocity, will gradually push the system towards the target in physical state space. To structure the latent space to be suitable for such a control law, we introduce the technique of pseudo-target labeling and the objective of latent space PD-control Lyapunov risk. 

    We test our method (ProCL) in two 2-D and one 3-D simulated environments. Under our evaluations, ProCL produces latent space where a PD-controller can be applied. It is shown that ProCL outperforms baseline method in both goal reaching and trajectory tracking tasks. 
    
    In summary, the main contribution of this work is to develop a framework that induces proportional derivative controllability on latent space. Our method is simple to implement, extends existing LCE methods, and shows superior performance in various environments.

\section{Preliminaries}
\label{sec:preliminaries}

\subsection{PD Control}

    Proportional derivative control (PD-control) is a class of feedback control~\cite{johnson2005pid,richard2008modern} which produces commands that are proportional to an error term and its derivative:
    \begin{equation}
        u_t = K_p e_t + K_d \frac{d}{dt} e_t
    \end{equation}

    Gain terms $(K_p, K_d)$ applied to error $e$ in proportion and its derivative respectively, shape the feedback response. PD control is widely applied in industry across numerous domains, providing a simple mechanism to stabilise systems. Like many other classical control methods, PD control usually is applied to relatively low-dimensional system with possibly known system model. In this work, we extend the applicability of PD control to the setting where only high dimensional visual input is provided.

\subsection{Lyapunov Stability}
    To motivate our design to enforce PD-controllability in latent space, we briefly introduce Lyapunov stability~\cite{haddad2011nonlinear,goldhirsch1987stability, park2007performance}. In our setting, we do not assume knowing a linear system model in priori. In fact, even if the underlying system is linear, the images and the embeded latent representation can very likely be nonlinear. Therefore, techniques in analysing linear system stability are no longer applicable. To characterize the stability of nonlinear systems, stability in the sense of Lyapunov is the most common way given the difficulty to derive a closed-form expression of a trajectory. The technique is to prove the existence of a Lyapunov function that is defined as follows: suppose that $x = 0$ is an equilibrium point for a dynamical system $\frac{dx}{dt} = f(x)$ where $f: \mathcal{X} \to \mathcal{R}^n$ is the system dynamics that is locally Lipschitz~\cite{haddad2011nonlinear}. Let $V: \mathcal{X} \to \mathcal{R}$ be a continuously differentiable function such that
    \begin{equation} \label{lyapunov_V}
        V(0) = 0 \mbox{,} \;\;\;\;  \mbox{and} \;\; \;\; V(x) > 0  \;\;\;\; \forall x \in \mathcal{X} \setminus \{0\}
    \end{equation}
    and the Lie derivative of $V$ along the trajectories
    \begin{equation} \label{lyapunov}
        L_f V = \frac{dV}{dt} = \nabla V(x)^T \frac{dx}{dt} = \nabla V(x)^T f(x) \leq 0
          \;\;\;\; \forall x \in \mathcal{X}
    \end{equation}
    then the origin is stable and $V$ is a Lyapunov function. 

\subsection{Predictive Coding, Consistency, Curvature}

    Predictive Coding, Consistency, Curvature (PC3)~\cite{pc3} is a recently proposed LCE framework that we  use as the backbone of our method. Under this framework, we jointly learn an encoder $E: \mathcal{X} \to \mathcal{Z}$ and latent space dynamics $F: \mathcal{Z} \times \mathcal{U} \to \mathcal{Z}$ to maximize next observation predictability without reconstruction. In practice, $\mathcal{X}$ is constructed as concatenation of multiple images to capture velocity. 
    
    The core idea is to use the following predictive coding loss to maximize the mutual information between $\mathcal{X}$ and $\mathcal{Z}$:
    \begin{equation} \label{cpc}
        L_{\mathrm{cpc}}(E,F) =
        - \mathbb{E} 
        \left[ 
            \frac{1}{K} \sum_i \mathrm{ln}
            \frac{F(E(x_{t+1}^{i}) | E(x_t^{i})), u_t^i)}
            {\frac{1}{K}\sum_j F(E(x_{t+1}^{i}) | E(x_t^{j})), u_t^j)} 
        \right]
    \end{equation}
    
    In addition to the CPC loss (\ref{cpc}), a consistency loss is also employed to enforce that a predicted latent embedding is consistent with the encoding from the next observation:
    \begin{equation}
        L_{\mathrm{cons}}(E,F) =
        - \mathbb{E}_{p(x_{t+1},x_t,u_t)} 
        \left[
            \mathrm{ln} F(E(x_{t+1})|E(x_t),u_t)
        \right]
    \end{equation}
    
    Finally, a curvature loss is used to encourage local-linearity. The curvature of $F$ is measured by by computing the first-order Taylor expansion error incurred when evaluating at $\tilde{z} = z+\eta_z$ and $\tilde{u} = u+\eta_u$,
    \begin{equation}
        \begin{split}
            L_{\mathrm{curv}}(F) &= 
                \mathbb{E}_{\eta \sim \mathcal{N}(0,\sigma{I})} 
                    \lVert 
                        f_{\mathcal{Z}}(\tilde{z}, \tilde{u}) -
                        (\nabla_z f_{\mathcal{Z}}(\tilde{z}, \tilde{u})\eta_z  \\ &\quad +
                        \nabla_u f_{\mathcal{Z}}(\tilde{z}, \tilde{u})\eta_u
                        ) -
                        f_{\mathcal{Z}}(z,u) 
                    \rVert
        \end{split}
    \end{equation}

    The complete PC3 loss is the combination of all three terms above:
    \begin{equation} \label{pc3}
    \begin{split} 
        L_{\mathrm{pc3}}(E,F) &= \lambda_1 L_{\mathrm{cpc}}(E,F)
                                +\lambda_2 L_{\mathrm{cons}}(E,F) \\
                              &\quad +\lambda_3 L_{\mathrm{curv}}(E,F)
    \end{split}
    \end{equation}

\section{Proportional derivative Controllable Latent Space}
\label{sec:methods}
    Consider the task of controlling dynamical systems in the form $s_{t+1} = f_S(s_t, u_t)$, $f_S$ is the system dynamics. We are interested in the case in which we can only access to high-dimensional visual observation $x_t$ and desire to drive the system to match a reference image in goal reaching task or an image sequence in trajectory tracking task. This scenario has wide applications in real-world when only vision input is provided. 
    
    \subsection{Motivation} \label{method_motivation}
    The motivation of our model is straightforward. We find that even though the underlying system is PD-controllable, it is difficult to perform PD-control on latent space learned by regular LCE method like PC3 as described above. The first challenge is that the correspondence between latent state dimension and control dimension is hard to recover. Consider a simple 2D point mass environment with 2D control. This generally is modeled with 4D latent embedding (2D for position and 2D for velocity). After the model is trained, we do not know which dimension is configuration (position in point mass environment) and which dimension is velocity. The correspondence between configuration and velocity can be solved by decoupling latent state $z$ in to configuration, velocity pair $(h,v)$ and have $h_t = h_{t-1} + \Delta t \cdot v_t $, but the correspondence between state and control is a harder problem. Local linearity is promoted using curvature loss, but it is still not easy to fully disentangle the corresponding pair of each degree of freedom of state and control. Manually viewing the generated latent map to verify it is disentangled and link each control dimension to each state dimension is possible, but is obviously not scalable to higher dimension environments. 
    
    Moreover, even if we manage to find a correspondence among configuration, velocity and control either by manual visual inspection or by imposing additional heuristics regulations~\cite{jaques2021newtonianvae}, there is no objective in the training procedure that drives PD-controllability. LCE methods only ensure that the latent encoding and the latent transition dynamics are predictive of next observation. This objective supports, e.g., planning algorithms on the latent space in a MPC framework, but does not encourage it to directly be PD-controllable.  
    
    To tackle above challenges, we propose the Proportional Derivative Controllable Latent (ProCL) method. ProCL extends PC3 to enable PD-control on latent space. First, we structure the latent state $z$ into a (configuration, velocity) pair $(h,v)$. This design can be easily implemented with simple modification to $E$ and $F$ in PC3 networks. More concretely, we refer to individual image frames at each timestep as $i_t$ and two consecutive frames $(i_t, i_{t-1})$ as $x_t$ to capture velocity. We have $z_t = (h_t, v_t) = E(x_t)$, where $h_t = e(i_t)$ and $v_t = (e(i_t) - e(i_{t-1})) / \Delta t$. Following this, we can have forward dynamics model $z_{t+1} = F(z_t, u_t) = (h_{t+1}, v_{t+1})$, where $h_{t+1} = h_t + v_{t+1}$ and $v_{t+1} = g(h_t, v_t, u_t)$. $e$ and $g$ are implemented as convolutional neural network (CNN) and multi-layer perceptron (MLP) respectively. As discussed earlier, this design would create a consistent configuration representation $h$ and its velocity $v$ in the latent space. Other than this modification to PC3 networks, we add additional objective to learning loss function which is going to be described in the following subsection. In execution, the PD-control on the latent space will take the form: 
    \begin{equation} \label{procl_control}
        u_t = K_p (h_t^{\eqtext{target}}-h_t) + K_d (v_t^{\eqtext{target}}-v_t),
    \end{equation}    
    where $h_t^{\eqtext{target}}$ is obtained by encoding provided target image into latent space; in trajectory tracking task, $v_t^{\eqtext{target}}$ is obtained by taking finite difference of the consecutive reference latent embedding and in goal reaching task, $v_t^{\eqtext{target}}$ is simply 0. In other words, we perform PD-control on the latent representation $h$ embedded from single image $i$. Meanwhile, as presented in the subsequent subsection, we consider $z=(h,v)$ as the full state when deriving loss for control stability.
    
\subsection{Enforce PD-controllability on latent space}
    \begin{figure}[h!] 
      \centering 
        \includegraphics[width=3in]{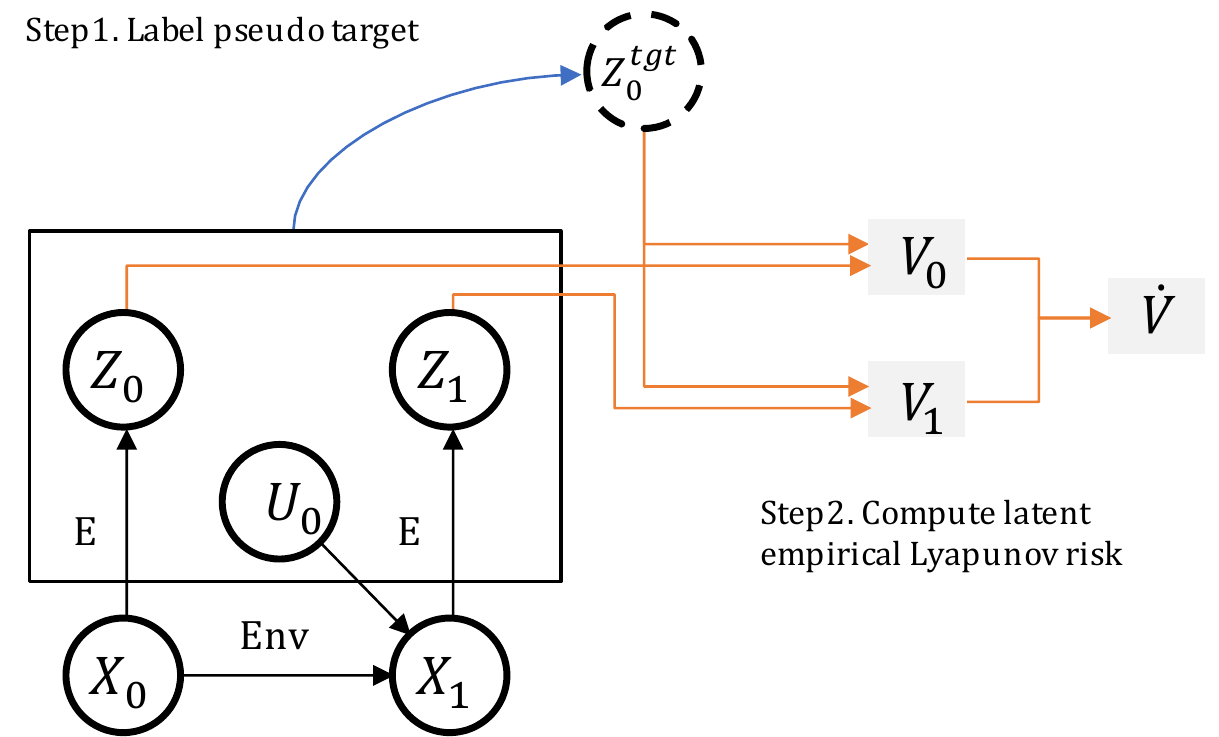} 
      \caption{Enforce PD-controllability by computing latent space PD-control Lyapunov risk as learning objective.}
      \label{method_fig}
    \end{figure}
    
    In this subsection, we introduce the technique to enforce PD-controllability in latent space. We add an additional objective so that a PD-controller would induce converging transitions that satisfying the Lyapunov condition in Eq.~\ref{lyapunov}.
    
    \textbf{Empirical Lyapunov risk:}  Following recent progress in combining Lyapunov stability with neural networks~\cite{almost_lyapunov, chang2020neural, boffi2020learning,mittal2020neural,ni4c}, we minimize additional loss function termed as empirical Lyapunov risk $R$. To introduce empirical Lyapunov risk, we define the Lyapunov function $V: \mathcal{Z} \to \mathcal{R}$ to be a continuously differentiable function. With the Lie derivative $L_f V$ along the controlled closed-loop dynamics $f$, the sample size $N$ and  the sampling distribution $\rho$, we have:
    \begin{equation}\label{risk}
        R = \frac{1}{N} \sum_{i=1}^{N} max(0, L_f V(z_i-z^{\eqtext{target}}))
    \end{equation}
    
    The empirical Lyapunov risk is non-negative, and when $V$ satisfies Eq.~\ref{lyapunov_V}-\ref{lyapunov}, it is a true Lyapunov function for $f$, i.e. this empirical risk reaches global minimum at $0$, for all sample size $N$ and distribution $\rho$. Given our setting that we only have access to transition tuples instead of the true dynamics $f$ of the system, we approximate the Lie derivative $L_f V$ along sample trajectories of the system through finite differences: 
    \begin{equation} 
        L_{f,\Delta t} V(z_t) = \frac{1}{\Delta t} (V(z_{t+1}-z^{\eqtext{target}}) - V(z_{t}-z^{\eqtext{target}})),
    \end{equation}
    where $z_t$ and $z_{t+1}$ are consecutive states and $\Delta t$ is the time difference between them. It's easy to see that $\textrm{lim}_{\Delta t \to 0} V(s) = L_{f,\Delta t} V(z_t)$. In principle $V$ can also be a trainable function of $z$. But for simplicity, we have $V = z^T Q z $ as a quadratic function of states with $Q$ as a tunable hyperparameter for this study. 
    
    After defining empirical Lyapunov risk as an additional objective to minimize, the immediate problem is how to effectively draw samples from a distribution following a PD-control law in latent space. We could iteratively collect samples by applying PD-control to some randomly selected target state, but that makes the training "on-policy" in the sense that transition tuple is only valid given the current latent encoding function. When the encoding function changes, the stored transition no longer follows proportional derivative control law in latent space, and thus such an approach is not sample efficient and may also be unstable in training.
    
    \textbf{Hindsight target labeling:} Inspired by Hindsight Experience Replay~\cite{andrychowicz2017hindsight,fang2018dher}, we propose to label a pseudo-target from a stored transition tuple $(z_t, u_t, z_{t+1})$. The intuition is that, given $z_t$, $u_t$ and the gain matrices $K_p$ and $K_d$, which are hyperparameters, we can invert this system to sample a series of targets $z^{\eqtext{target}}_{t}$ that reconciles with $z_t$, $u_t$ and the gain matrices. That is {\em we compute the encoding $z^{\eqtext{target}}_{t}$ that would explain the control $u_t$ we take at $z_t$.} As such, this is effectively "off-policy" in the sense that we can use all transition tuples collected by any random policy to update the embedding function. With $z_t = (h_t, v_t)$, we first sample $v_t^{\eqtext{target}} \in \textrm{Prior}(v)$, where $\textrm{Prior}(v)$ is implemented as a buffer of all $v_t$ in the current mini-batch. And then we can treat $h_t^{\eqtext{target}}$ as the only unknown variable in solving Eq.~\ref{procl_control} to get the following expression:
    \begin{equation} \label{h_target}
        h_t^{\eqtext{target}}  = K_p^{-1} (u_t + K_p h_t - K_d (v_t^{\eqtext{target}} - v_t))
    \end{equation}
    
    $z_t^{\eqtext{target}}$ is then naturally composed as $(h_t^{\eqtext{target}}, v_t^{\eqtext{target}})$. 
    
    \textbf{Latent space PD-control Lyapunov risk:} With hindsight pseudo-target labeling, we now have an expanded transition tuple $(z_t, u_t, z_{t+1}, z_t^{\eqtext{target}})$, on top of which we can then express the approximated Lie derivative as:
    \begin{equation} \label{latent_P_risk}
        L_{f,\Delta t} V(z_t) = \frac{1}{\Delta t} (V(z_{t+1} - z_t^{\eqtext{target}}) - V(z_{t} - z_t^{\eqtext{target}})),
    \end{equation}
    which is then used to replace $L_f V$ in Eq.~\ref{risk} to compute the empirical Lyapunov risk $R$. Notice that given a replay buffer of transitions pre-collected by a random policy, $R$ is now a function of latent space encoding function $E$. Minimizing $R$ is then to find function $E$ so that change in $V$ brought by PD-controller in latent space encoded by $E$ is non-positive. We thus denote the latent space PD-control Lyapunov risk as $R(E)$.
    
    We combine this objective $R(E)$ with the original PC3 loss function (\ref{pc3}). Our overall objective is thus:
    \begin{equation} \label{overall_loss}
        L_{\mathrm{ProCL}}(E,F) = \lambda_{pc3} L_{pc3}(E,F) + \lambda_{R} R(E)
    \end{equation}
    
    To summarize, as shown in Fig.~\ref{method_fig}, we minimize the empirical Lyapunov risk evaluated for PD-controller on latent space as an objective to enforce latent space is PD-controllable. Hindsight pseudo-target labeling is employed to efficiently provide training examples to perform training. Pseudo-code~\ref{code} is also shown below. 
    
    \begin{algorithm} \label{code}
    \caption{Proportional Derivative Controllable Latent Space}\label{code}
    \textbf{Input:} PD-control gain $K_p$, $K_d$ and Lyapunov function V
    \begin{algorithmic}[1]
    \State Collect experience using random control and store into replay buffer $B$
    \State Initialize encoding function $E$ and dynamics model $F$
    \While {not finished}: 
        \State Sample $x_t, u_t, x_{t+1}$ from $B$
        \State Sample $v_t^{\eqtext{target}}$ from $\textrm{Prior}(v_t)$
        \State Compute PC3 loss using Eq.~\ref{pc3}
        \State Compute $h_t^{\eqtext{target}}$ using Eq.~\ref{h_target}
        \State Compute latent PD-control Lyapunov risk using Eq.~\ref{latent_P_risk} and Eq.~\ref{risk}
        \State Update $E$ and $F$ with the objective defined in Eq.~\ref{overall_loss}
    \EndWhile
    \end{algorithmic}
    \end{algorithm}

\subsection{More considerations} \label{noise}
\label{sec:experiment}
\begin{figure*}[h!] 
  \centering 
    \includegraphics[width=6in]{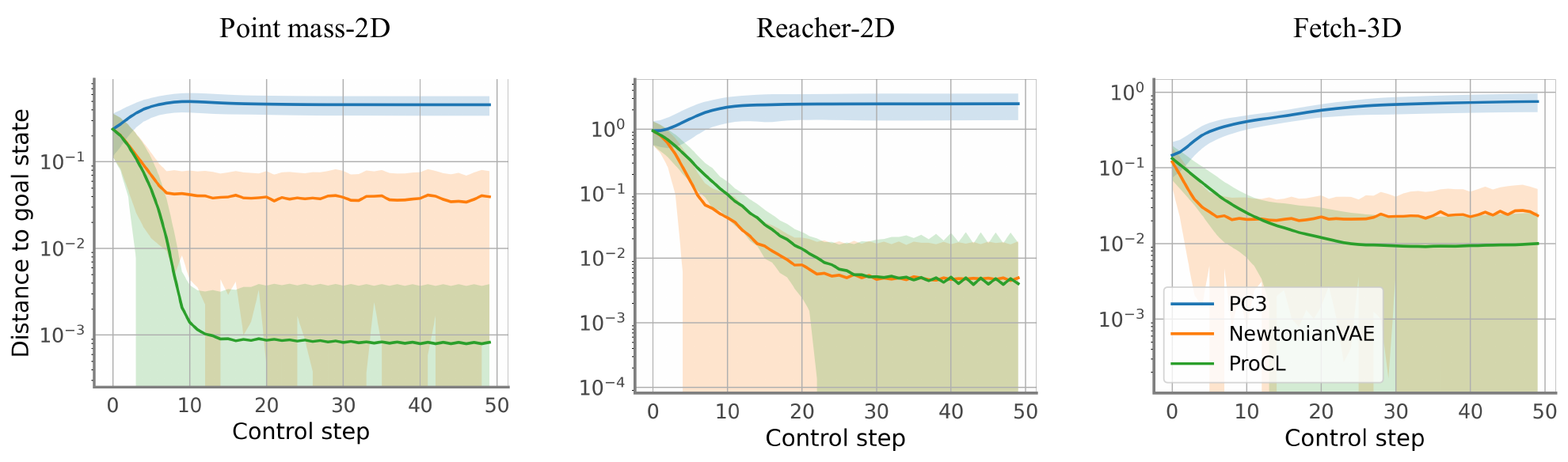}
  \caption{Goal reaching performance using PD controller on PC3 and ProCL and PID controller on NewtonianVAE. The y axes are in log-scale. Shaded area represents plus or minus one standard deviation over 50 episodes.}
  \label{ablation}
\end{figure*}
One issue applying Lyapunov risk $R$ as training objective on latent space is that $R$ is sensitive to non-volume preserving transformations of the latent space. That is shrinking the overall size of latent map will minimize the value of $R$. This is similar to $L_{\mathrm{cons}}$ in PC3 training. Therefore, following~\cite{pc3}, we add Gaussian noise $\epsilon \in \mathcal{N}(0,\sigma^2 I)$ to the next state encoding $E(x_{t+1})$. Since the noise variance is fixed, $L_{\mathrm{cpc}}$ can maximized by expanding the latent space, thus balancing the contraction behaviour produced by $R$ and $L_{\mathrm{cons}}$. 

\section{Experiments}

\begin{figure}[h!]
  \centering 
  \subfigure[Point mass-2D]{
    \includegraphics[width=1.in]{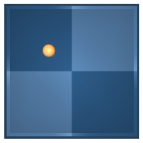}
  }
  \subfigure[Reacher-2D]{
    \includegraphics[width=1.in]{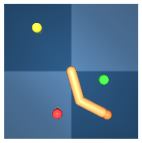}
  }
  \subfigure[Fetch-3D]{
    \includegraphics[width=1.in]{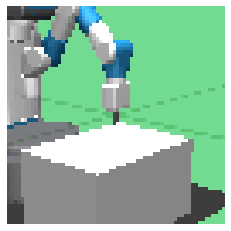}
  }
  \caption{Environment observation samples: Point mass-2D and Reacher-2D and Fetch-3D.}
  \label{fig:envs}
\end{figure}

In this section, we report evaluations of our method ProCL, along with comparison with baseline method. The experiments are based on three image-based environments: Point mass, Reacher-2D and Fetch-3D. Observation examples can be found in Fig.~\ref{fig:envs}.

\textbf{Point mass-2D:} The Point mass-2D environment is adapted from PointMass system from deepmind control suite~\cite{tassa2018deepmind}. The mass has two degree of freedom and is linearly actuated by 2D control to move on a plane. The movement is bounded by the frame edges.

\textbf{Reacher-2D:} The Reacher-2D environment is adapted from the Reacher environment in deepmind control suite. It has two degree of freedom in configuration and fully actuated by 2D control. Following~\cite{jaques2021newtonianvae}, we limit robot’s middle joint angle so that only it bends in one direction and also limit the origin joint angle range to between -160 and 160 degrees to avoid discontinuity in full circular motion. 

\textbf{Fetch-3D:} The Fetch-3D environment is adapted from FetchReach-V1 environment from OpenAI Gym~\cite{1606.01540}. No adjustment is made other than that we resize the pixel observation to 128 by 128 and then crop the center 64 by 64 of it. The robot arm is actuated in the 3D end-effector position with 3D control as acceleration of the end-effector position. This environment is more challenging due to the 3D visual scene and partial occlusions came with it.

\textbf{Training data generation:} The training data is collected using random control interacting with environments. We generate 10000 time-steps' transitions for Point mass-2D and Reacher-2D, and 50000 time-steps' transitions for Fetch-3D.

\textbf{Baseline methods:} We compare our model to two baselines, the original PC3~\cite{pc3} with PD control applied to learned latent space and Newtonian VAE~\cite{jaques2021newtonianvae} with PID control. For both our method ProCL and PC3, We use the same network architecture described in PC3 and apply the change in networks described in the end of section~\ref{method_motivation}. For NewtonianVAE, we thankfully obtained the original implementation from the authors. We empirically find that PD and PID control generate similar performances for NewtonianVAE and thus choose to directly compare with PID control with the original hyperparameters shown in the paper. No additional modification is made. 

\begin{figure*}[h!] 
  \centering 
    \includegraphics[width=6.5in]{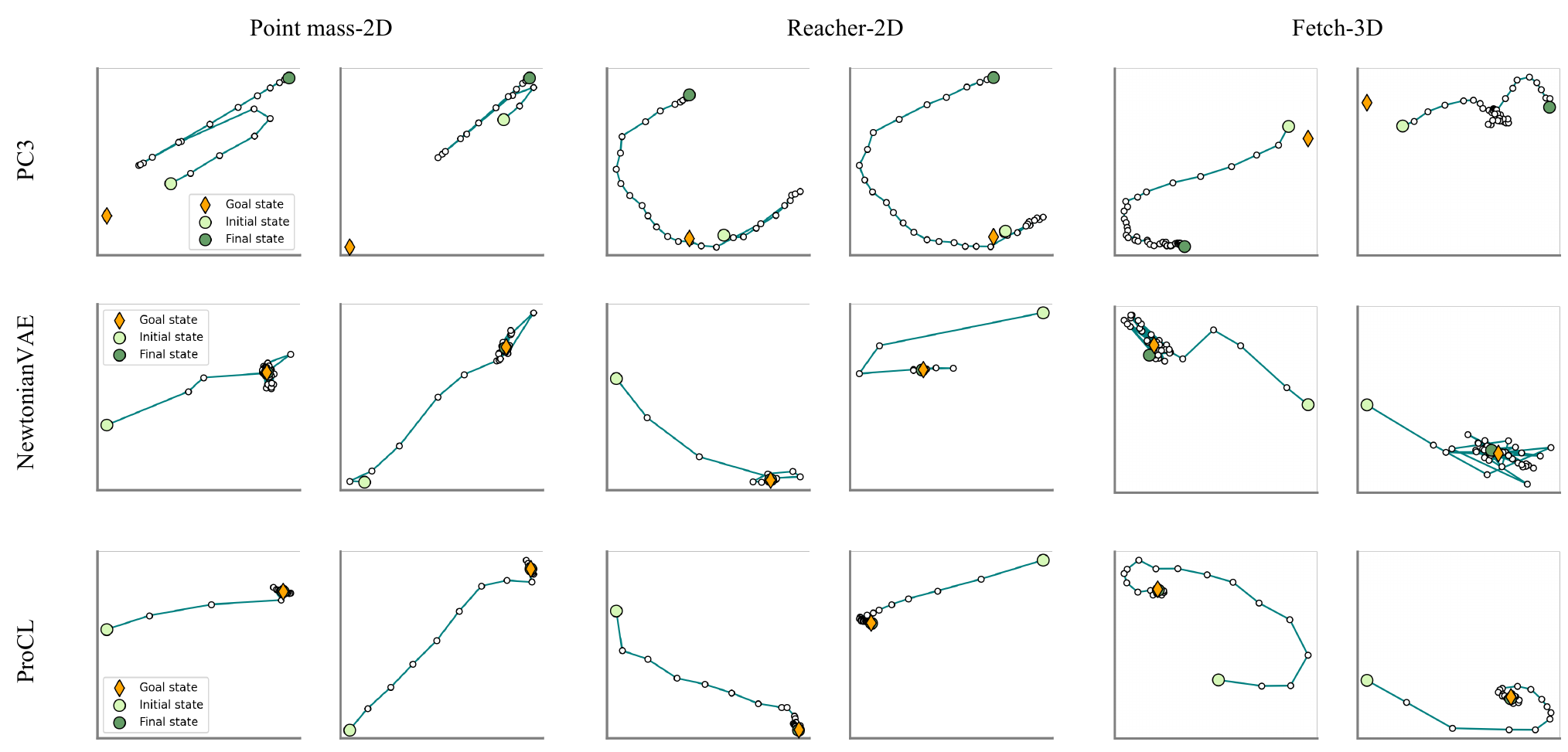}
  \caption{Goal reaching trajectories are shown in the learned latent state. We can see that across all three environments, ProCL learns a latent space that is compatible with PD-controller to produce robust reaching from start state to target state.}
  \label{goal_reaching_trajectory}
\end{figure*}
\subsection{Visual goal reaching}

In this sub-section, we compare the PD-controllability properties of the latent space generated by ProCL and the baseline methods on all three simulated environments. We train both ProCL and PC3 using the pre-generated dataset to learn embedding functions. During test time, we use PD-controller to control the system to reach a randomly sampled target state. For NewtonianVAE, we obtain the pretrained model on the same environments kindly shared by the author. Following the original implementation, PID control ($K_p=8\cdot I$, $K_i=2\cdot I$, $K_d=0.5\cdot I$) is applied to the learned latent state by NewtonianVAE. 

Trajectories shown in Fig.~\ref{goal_reaching_trajectory} demonstrate the advantage of our method's PD-controllability in the latent space. PD-controller applied to ProCL's latent space can generate converging dynamics towards target. NewtonianVAE can also achieves converging behaviour in latent space but with more oscillation near the target. Meanwhile, PC3's latent space does not have such property: the PD-controlled trajectory either largely oscillates around target or can not approach in the right direction across the whole time period.

\begin{figure*}[h!] 
  \centering 
    \includegraphics[width=6in]{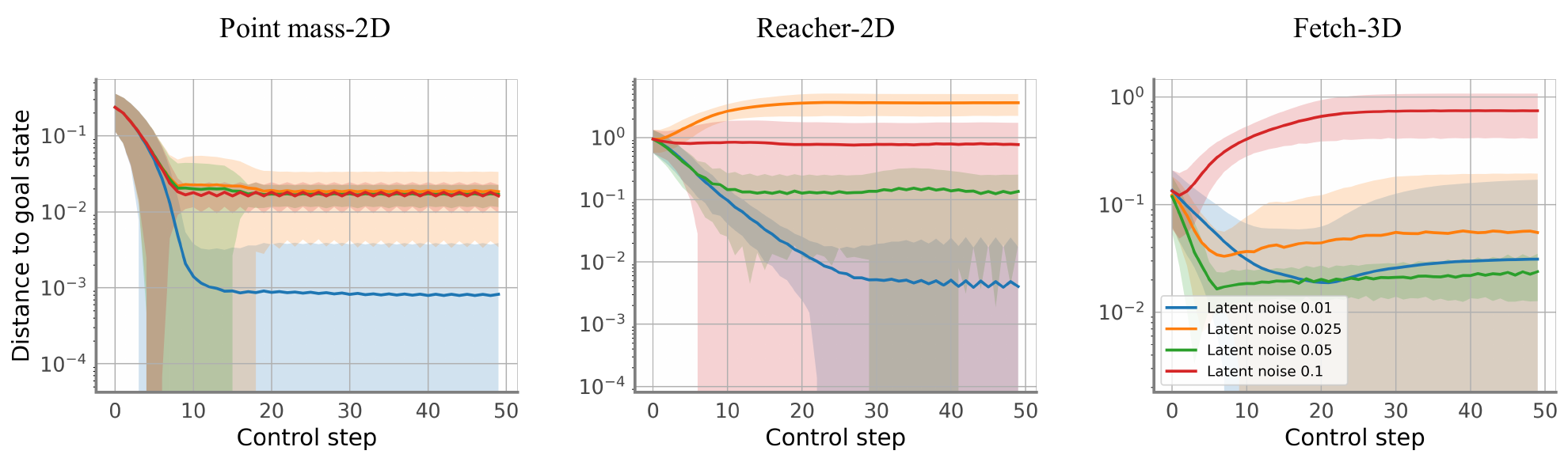}
  \caption{Ablation study on hyperparameter. Convergence rates using different latent noise during the training of ProCL. The y axes are in log-scale. Shaded area represents plus or minus one standard deviation over 50 episodes.}
  \label{goal_reaching_distance}
\end{figure*}

To more closely comparing the performance, we present the convergence curves of the L2 distance in the ground truth state configuration space in Fig.~\ref{goal_reaching_distance}. We can see that the average distance with our method converges quickly to close to zero. ProCL outperforms NewtonianVAE in two of three environments. PD-controller on PC3 does not make progress across many steps.

\textbf{Hyperparameters:} For all experiments, we set $\lambda_{pc3}, \lambda_R, \lambda_1, \lambda_2, \lambda_3$ to 1, 10, 1, 1, 10 respectively. We also have fixed PD-control gain matrix $K_p$ as $10\cdot I$ and $K_d$ as $2\cdot I$. Quadratic Lyapunov function matrix $Q_h$ is set to $I$ and $Q_v$ is set to $0.1\cdot I$. $\Delta t$ is set to match environment timestep for point mass-2D and reacher-2d, and is set to 1 for fetch-3D. The only hyperparameter we tune is the noise $\epsilon$ added to latent embedding during training of ProCL. The effect of PD-control is related to the scale of the latent space which is balanced by the noise $\epsilon$ as discussed in section~\ref{noise}. We therefore tune the latent noise $\epsilon$ as a way to control the scale of the latent map, which then relates to the scale of the control. As shown in Fig.~\ref{ablation}, we present ablation study using different noise level. In our experiments, $\epsilon=0.01$ generates the best performance across all three environments and is thus chosen for both goal reaching and trajectory tracking task.

\subsection{Visual trajectory tracking}

\begin{figure*}[h!] 
  \centering 
      \subfigure{
        \includegraphics[width=4.25in]{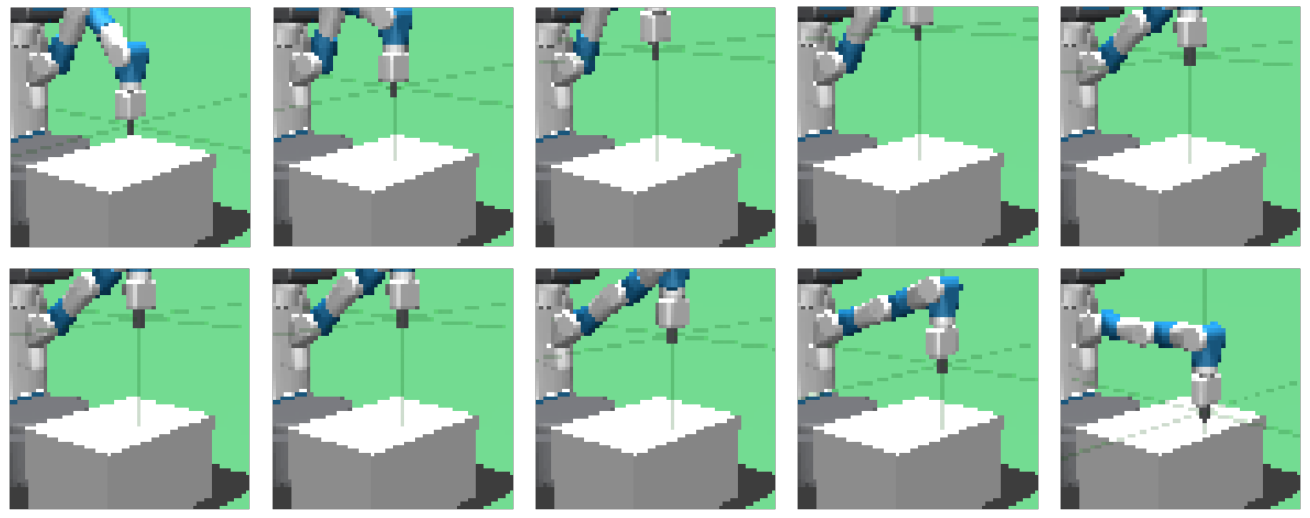}
        }
      \subfigure{
        \includegraphics[width=1.65in]{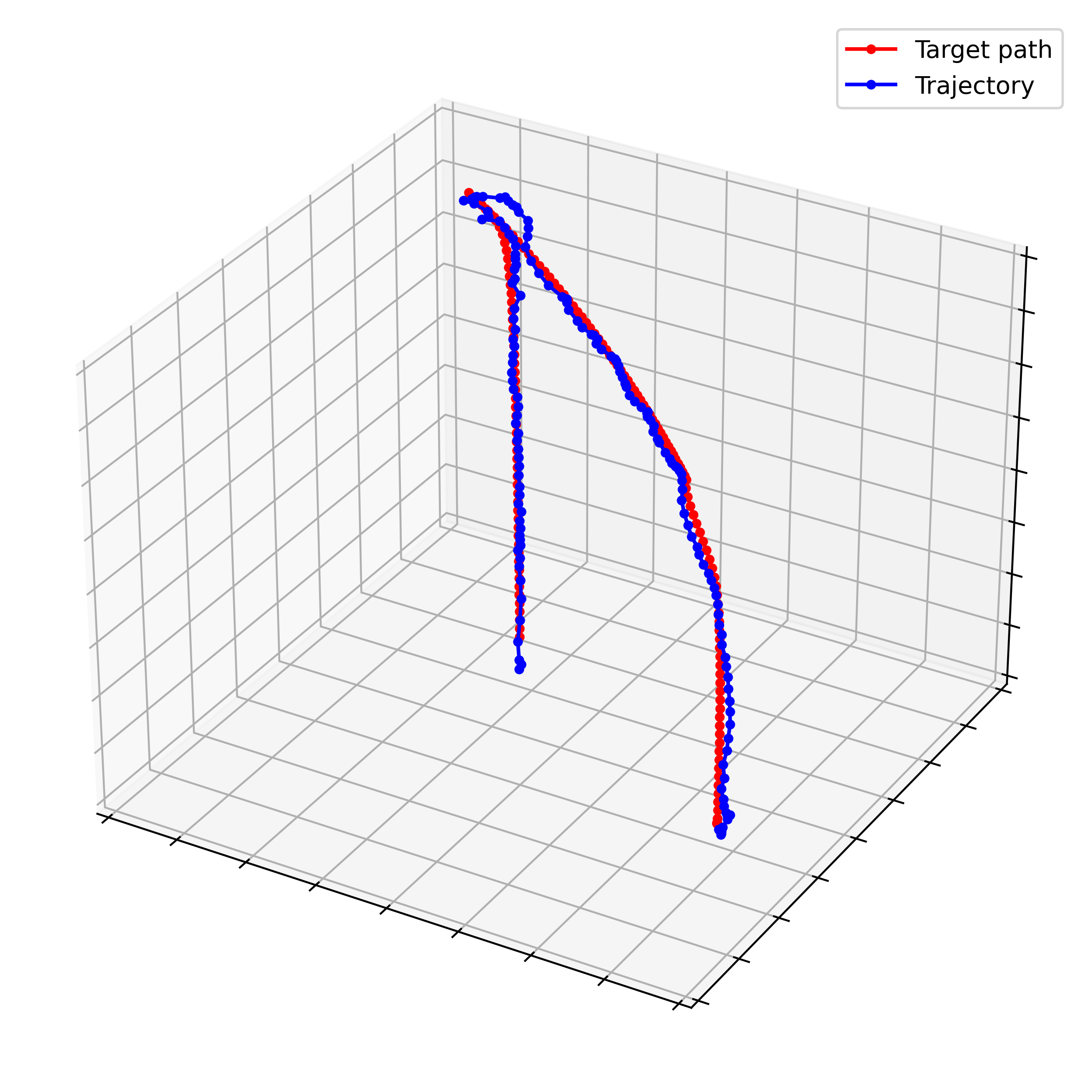}
        }
    \caption{ Performance in trajectory tracking.
        \textbf{Left:} ProCL's observed frames during visual trajectory tracking. \textbf{Right:} 3D view of target path and tracking trajectory by ProCL.}
  \label{tracking_traj}
\end{figure*}

Besides goal reaching, we also show how ProCL combined with PD-control alone can enable visual input trajectory tracking, using the Fetch-3D environment. Contrary to NewtonianVAE~\cite{jaques2021newtonianvae} where dynamic movement primitives (DMP)~\cite{ijspeert2013dynamical, schaal2006dynamic} needs to be fitted with demonstrated data, tracking with PD-control with Eq.~\ref{procl_control} can be performed directly using our approach. We generate a sequence of demonstration trajectory in Fetch-3D environment and then use the learned latent encoding function to embed the visual demonstrations into latent space trajectory. PD-controller is then applied to track this latent trajectory in execution with the same hyperparameters ($K_p=10\cdot I$, $K_d=2\cdot I$) as in the previous subsection. As shown in Fig.~\ref{tracking_traj}, our method is able to closely track the demonstrated visual rollout. 

\section{Related Work}
\label{sec:related_work}
Modeling high-dimensional visual inputs with latent encoding has been of interest for a long time. Variational autoencoder (VAE)~\cite{kingma2013auto, doersch2016tutorial} provides a general framework to conduct variational inference of latent variable over visual inputs. Following this direction, it has been shown success in training a latent dynamics model in the latent space of VAE to perform control tasks~\cite{ha2018world, kobayashis2020q, meo2021multimodal} as well as perform video prediction and generation~\cite{pu2016variational, he2018probabilistic}.

In the space of modeling visual observation dynamics for control, learning Controllable Embedding (LCE) is a recent framework to jointly learn embedding function from the high-dimensional visual input into a low-dimensional latent space, and a transition dynamics model in the latent space~\cite{watter2015embed, finn2016deep, zhang2019solar, banijamali2018robust, pc3, Levine2020Prediction}. There have been heuristics to guide the learning of meaningful latent state. E2C~\cite{watter2015embed} uses locally linear neural network to parameterize the latent transition dynamics. Some other methods such as PCC~\cite{Levine2020Prediction} and PC3~\cite{pc3} also desire local linearity in latent space but choose to employ general neural network with curvature as additional penalty. Next observation prediction has been widely used as training signal to learn meaningful representation in most LCE methods like E2C and PCC. Recently proposed PC3 uses
predictive coding as an alternative to minimizes predictive suboptimality, which bypasses the need for a decoder network.

As we discussed earlier, LCE methods such as E2C, PCC and PC3 usually apply MPC methods to perform control in execution, of which the intensive computational cost becomes the runtime bottleneck. This limitation also applies to some of more recent targeting more complex environment such as PlaNet~\cite{hafner2019learning} which uses Cross Entropy Method (CEM)~\cite{kobilarov2012cross,de2005tutorial} to search for the best action sequences under learned model. One alternative is to not separate control from latent space encoding and dynamics learning. SOLAR~\cite{zhang2019solar}, for example, performs policy optimization alongside with model learning. The learned latent space is thus becoming task-specific instead of dynamics-specific as in the previous approaches . Other recent work such as Dreamer~\cite{hafner2019dream} and MuZero~\cite{schrittwieser2020mastering} also predicts the task specific reward and value function in the learned latent dynamics model to further improve the performance.

Introduction of physical knowledge prior~\cite{abraham2017model,jaques2019physics,jonschkowski2017pves,li2019learning, toth2019hamiltonian} has also shown to enhance the modeling and control of latent dynamics. It's presented in~\cite{zhong2020unsupervised} that a Lagrangian dynamics can be learned from images to improve dynamic prediction and generalization. Among all the works, Newtonian VAE~\cite{jaques2021newtonianvae} is perhaps the most relevant work to ours. Their motivation is to regularize the latent space embedding along each dimension of control with the heuristics of each dimension of control applies strictly positive change to corresponding dimension in the latent state. Although similar to our approach in applying PD-controller to latent space, we argue that their method does not have a principled objective to optimize for such PD-controllability. In contrary, our approach directly minimize the latent space PD-control Lyapunov risk to enforce PD-controllability.

\section{Conclusion}
\label{sec:conclusion}
We introduce proportional derivative controllable latent space (ProCL) model to learn PD-controllable latent space from visual perceptions. With the assumption that the underlying physical system is PD-controllable, ProCL can enable PD-control on learned latent space which bypasses the computational burden for MPC type of method in execution time. Experimental results show that our method can produce robust and superior performance in both goal reaching and visual trajectory tracking tasks across multiple environments.

\addtolength{\textheight}{-2.5cm}  
\bibliographystyle{IEEEtran}
\bibliography{IEEEabrv,bbl}

\begin{thebibliography}{10}
\providecommand{\url}[1]{#1}
\csname url@rmstyle\endcsname
\providecommand{\newblock}{\relax}
\providecommand{\bibinfo}[2]{#2}
\providecommand\BIBentrySTDinterwordspacing{\spaceskip=0pt\relax}
\providecommand\BIBentryALTinterwordstretchfactor{4}
\providecommand\BIBentryALTinterwordspacing{\spaceskip=\fontdimen2\font plus
\BIBentryALTinterwordstretchfactor\fontdimen3\font minus
  \fontdimen4\font\relax}
\providecommand\BIBforeignlanguage[2]{{%
\expandafter\ifx\csname l@#1\endcsname\relax
\typeout{** WARNING: IEEEtran.bst: No hyphenation pattern has been}%
\typeout{** loaded for the language `#1'. Using the pattern for}%
\typeout{** the default language instead.}%
\else
\language=\csname l@#1\endcsname
\fi
#2}}

\bibitem{watter2015embed}
M.~Watter, J.~T. Springenberg, J.~Boedecker, and M.~Riedmiller, ``Embed to
  control: A locally linear latent dynamics model for control from raw
  images,'' 2015.

\bibitem{finn2016deep}
C.~Finn, X.~Y. Tan, Y.~Duan, T.~Darrell, S.~Levine, and P.~Abbeel, ``Deep
  spatial autoencoders for visuomotor learning,'' in \emph{2016 IEEE
  International Conference on Robotics and Automation (ICRA)}.\hskip 1em plus
  0.5em minus 0.4em\relax IEEE, 2016, pp. 512--519.

\bibitem{zhang2019solar}
M.~Zhang, S.~Vikram, L.~Smith, P.~Abbeel, M.~J. Johnson, and S.~Levine,
  ``Solar: Deep structured representations for model-based reinforcement
  learning,'' 2019.

\bibitem{banijamali2018robust}
E.~Banijamali, R.~Shu, M.~Ghavamzadeh, H.~Bui, and A.~Ghodsi, ``Robust
  locally-linear controllable embedding,'' 2018.

\bibitem{pc3}
R.~Shu, T.~Nguyen, Y.~Chow, T.~Pham, K.~Than, M.~Ghavamzadeh, S.~Ermon, and
  H.~Bui, ``Predictive coding for locally-linear control,'' in
  \emph{International Conference on Machine Learning}.\hskip 1em plus 0.5em
  minus 0.4em\relax PMLR, 2020, pp. 8862--8871.

\bibitem{Levine2020Prediction}
N.~Levine, Y.~Chow, R.~Shu, A.~Li, M.~Ghavamzadeh, and H.~Bui, ``Prediction,
  consistency, curvature: Representation learning for locally-linear control,''
  2020.

\bibitem{camacho2013model}
E.~F. Camacho and C.~B. Alba, \emph{Model predictive control}.\hskip 1em plus
  0.5em minus 0.4em\relax Springer science \& business media, 2013.

\bibitem{levine2014learning}
S.~Levine and P.~Abbeel, ``Learning neural network policies with guided policy
  search under unknown dynamics.'' in \emph{NIPS}, vol.~27.\hskip 1em plus
  0.5em minus 0.4em\relax Citeseer, 2014, pp. 1071--1079.

\bibitem{hirose2019deep}
N.~Hirose, F.~Xia, R.~Mart{\'\i}n-Mart{\'\i}n, A.~Sadeghian, and S.~Savarese,
  ``Deep visual mpc-policy learning for navigation,'' \emph{IEEE Robotics and
  Automation Letters}, vol.~4, no.~4, pp. 3184--3191, 2019.

\bibitem{finn2017deep}
C.~Finn and S.~Levine, ``Deep visual foresight for planning robot motion,'' in
  \emph{2017 IEEE International Conference on Robotics and Automation
  (ICRA)}.\hskip 1em plus 0.5em minus 0.4em\relax IEEE, 2017, pp. 2786--2793.

\bibitem{lynch2020learning}
C.~Lynch, M.~Khansari, T.~Xiao, V.~Kumar, J.~Tompson, S.~Levine, and
  P.~Sermanet, ``Learning latent plans from play,'' in \emph{Conference on
  Robot Learning}.\hskip 1em plus 0.5em minus 0.4em\relax PMLR, 2020, pp.
  1113--1132.

\bibitem{johnson2005pid}
M.~A. Johnson and M.~H. Moradi, \emph{PID control}.\hskip 1em plus 0.5em minus
  0.4em\relax Springer, 2005.

\bibitem{richard2008modern}
R.~C. Dorf and R.~H. Bishop, \emph{Modern control systems}.\hskip 1em plus
  0.5em minus 0.4em\relax Pearson Prentice Hall, 2008.

\bibitem{haddad2011nonlinear}
W.~M. Haddad and V.~Chellaboina, \emph{Nonlinear dynamical systems and
  control}.\hskip 1em plus 0.5em minus 0.4em\relax Princeton university press,
  2011.

\bibitem{goldhirsch1987stability}
I.~Goldhirsch, P.-L. Sulem, and S.~A. Orszag, ``Stability and lyapunov
  stability of dynamical systems: A differential approach and a numerical
  method,'' \emph{Physica D: Nonlinear Phenomena}, vol.~27, no.~3, pp.
  311--337, 1987.

\bibitem{park2007performance}
S.~Park, J.~Deyst, and J.~P. How, ``Performance and lyapunov stability of a
  nonlinear path following guidance method,'' \emph{Journal of guidance,
  control, and dynamics}, vol.~30, no.~6, pp. 1718--1728, 2007.

\bibitem{jaques2021newtonianvae}
M.~Jaques, M.~Burke, and T.~Hospedales, ``Newtonianvae: Proportional control
  and goal identification from pixels via physical latent spaces,'' 2021.

\bibitem{almost_lyapunov}
Y.-C. Chang and S.~Gao, ``Stabilizing neural control using self-learned almost
  lyapunov critics,'' in \emph{ICRA}, 2021.

\bibitem{chang2020neural}
Y.-C. Chang, N.~Roohi, and S.~Gao, ``Neural lyapunov control,'' \emph{arXiv
  preprint arXiv:2005.00611}, 2020.

\bibitem{boffi2020learning}
N.~M. Boffi, S.~Tu, N.~Matni, J.-J.~E. Slotine, and V.~Sindhwani, ``Learning
  stability certificates from data,'' \emph{arXiv preprint arXiv:2008.05952},
  2020.

\bibitem{mittal2020neural}
M.~Mittal, M.~Gallieri, A.~Quaglino, S.~S.~M. Salehian, and J.~Koutn{\'\i}k,
  ``Neural lyapunov model predictive control,'' \emph{arXiv preprint
  arXiv:2002.10451}, 2020.

\bibitem{ni4c}
P.~Saha, M.~Egerstedt, and S.~Mukhopadhyay, ``Neural identification for
  control,'' \emph{IEEE Robotics and Automation Letters}, vol.~6, no.~3, pp.
  4648--4655, 2021.

\bibitem{andrychowicz2017hindsight}
M.~Andrychowicz, F.~Wolski, A.~Ray, J.~Schneider, R.~Fong, P.~Welinder,
  B.~McGrew, J.~Tobin, P.~Abbeel, and W.~Zaremba, ``Hindsight experience
  replay,'' \emph{arXiv preprint arXiv:1707.01495}, 2017.

\bibitem{fang2018dher}
M.~Fang, C.~Zhou, B.~Shi, B.~Gong, J.~Xu, and T.~Zhang, ``Dher: Hindsight
  experience replay for dynamic goals,'' in \emph{International Conference on
  Learning Representations}, 2018.

\bibitem{tassa2018deepmind}
Y.~Tassa, Y.~Doron, A.~Muldal, T.~Erez, Y.~Li, D.~de~Las~Casas, D.~Budden,
  A.~Abdolmaleki, J.~Merel, A.~Lefrancq, T.~Lillicrap, and M.~Riedmiller,
  ``Deepmind control suite,'' 2018.

\bibitem{1606.01540}
G.~Brockman, V.~Cheung, L.~Pettersson, J.~Schneider, J.~Schulman, J.~Tang, and
  W.~Zaremba, ``Openai gym,'' 2016.

\bibitem{ijspeert2013dynamical}
A.~J. Ijspeert, J.~Nakanishi, H.~Hoffmann, P.~Pastor, and S.~Schaal,
  ``Dynamical movement primitives: learning attractor models for motor
  behaviors,'' \emph{Neural computation}, vol.~25, no.~2, pp. 328--373, 2013.

\bibitem{schaal2006dynamic}
S.~Schaal, ``Dynamic movement primitives-a framework for motor control in
  humans and humanoid robotics,'' in \emph{Adaptive motion of animals and
  machines}.\hskip 1em plus 0.5em minus 0.4em\relax Springer, 2006, pp.
  261--280.

\bibitem{kingma2013auto}
D.~P. Kingma and M.~Welling, ``Auto-encoding variational bayes,'' \emph{arXiv
  preprint arXiv:1312.6114}, 2013.

\bibitem{doersch2016tutorial}
C.~Doersch, ``Tutorial on variational autoencoders,'' \emph{arXiv preprint
  arXiv:1606.05908}, 2016.

\bibitem{ha2018world}
D.~Ha and J.~Schmidhuber, ``World models,'' \emph{arXiv preprint
  arXiv:1803.10122}, 2018.

\bibitem{kobayashis2020q}
T.~Kobayashis, ``q-vae for disentangled representation learning and latent
  dynamical systems,'' \emph{IEEE Robotics and Automation Letters}, vol.~5,
  no.~4, pp. 5669--5676, 2020.

\bibitem{meo2021multimodal}
C.~Meo and P.~Lanillos, ``Multimodal vae active inference controller,''
  \emph{arXiv preprint arXiv:2103.04412}, 2021.

\bibitem{pu2016variational}
Y.~Pu, Z.~Gan, R.~Henao, X.~Yuan, C.~Li, A.~Stevens, and L.~Carin,
  ``Variational autoencoder for deep learning of images, labels and captions,''
  \emph{Advances in neural information processing systems}, vol.~29, pp.
  2352--2360, 2016.

\bibitem{he2018probabilistic}
J.~He, A.~Lehrmann, J.~Marino, G.~Mori, and L.~Sigal, ``Probabilistic video
  generation using holistic attribute control,'' in \emph{Proceedings of the
  European Conference on Computer Vision (ECCV)}, 2018, pp. 452--467.

\bibitem{hafner2019learning}
D.~Hafner, T.~Lillicrap, I.~Fischer, R.~Villegas, D.~Ha, H.~Lee, and
  J.~Davidson, ``Learning latent dynamics for planning from pixels,'' in
  \emph{International Conference on Machine Learning}.\hskip 1em plus 0.5em
  minus 0.4em\relax PMLR, 2019, pp. 2555--2565.

\bibitem{kobilarov2012cross}
M.~Kobilarov, ``Cross-entropy motion planning,'' \emph{The International
  Journal of Robotics Research}, vol.~31, no.~7, pp. 855--871, 2012.

\bibitem{de2005tutorial}
P.-T. De~Boer, D.~P. Kroese, S.~Mannor, and R.~Y. Rubinstein, ``A tutorial on
  the cross-entropy method,'' \emph{Annals of operations research}, vol. 134,
  no.~1, pp. 19--67, 2005.

\bibitem{hafner2019dream}
D.~Hafner, T.~Lillicrap, J.~Ba, and M.~Norouzi, ``Dream to control: Learning
  behaviors by latent imagination,'' \emph{arXiv preprint arXiv:1912.01603},
  2019.

\bibitem{schrittwieser2020mastering}
J.~Schrittwieser, I.~Antonoglou, T.~Hubert, K.~Simonyan, L.~Sifre, S.~Schmitt,
  A.~Guez, E.~Lockhart, D.~Hassabis, T.~Graepel, \emph{et~al.}, ``Mastering
  atari, go, chess and shogi by planning with a learned model,'' \emph{Nature},
  vol. 588, no. 7839, pp. 604--609, 2020.

\bibitem{abraham2017model}
I.~Abraham, G.~De~La~Torre, and T.~D. Murphey, ``Model-based control using
  koopman operators,'' \emph{arXiv preprint arXiv:1709.01568}, 2017.

\bibitem{jaques2019physics}
M.~Jaques, M.~Burke, and T.~Hospedales, ``Physics-as-inverse-graphics:
  Unsupervised physical parameter estimation from video,'' \emph{arXiv preprint
  arXiv:1905.11169}, 2019.

\bibitem{jonschkowski2017pves}
R.~Jonschkowski, R.~Hafner, J.~Scholz, and M.~Riedmiller, ``Pves:
  Position-velocity encoders for unsupervised learning of structured state
  representations,'' \emph{arXiv preprint arXiv:1705.09805}, 2017.

\bibitem{li2019learning}
Y.~Li, H.~He, J.~Wu, D.~Katabi, and A.~Torralba, ``Learning compositional
  koopman operators for model-based control,'' \emph{arXiv preprint
  arXiv:1910.08264}, 2019.

\bibitem{toth2019hamiltonian}
P.~Toth, D.~J. Rezende, A.~Jaegle, S.~Racani{\`e}re, A.~Botev, and I.~Higgins,
  ``Hamiltonian generative networks,'' \emph{arXiv preprint arXiv:1909.13789},
  2019.

\bibitem{zhong2020unsupervised}
Y.~D. Zhong and N.~Leonard, ``Unsupervised learning of lagrangian dynamics from
  images for prediction and control,'' \emph{Advances in Neural Information
  Processing Systems}, vol.~33, 2020.

\end{thebibliography}
\addtolength{\textheight}{-6cm}  








\end{document}